\documentclass[conference]{IEEEtran}

\usepackage{arxiv}
\usepackage[T1]{fontenc}    
\usepackage[utf8]{inputenc} 
\usepackage{hyperref}       
\usepackage{url}            

\usepackage{amsfonts}       
\usepackage{nicefrac}       
\usepackage{microtype}      

\usepackage{multirow}
\usepackage{booktabs}       
\usepackage{graphicx}       
\usepackage{adjustbox}
\usepackage{stfloats}
\usepackage{doi}
\usepackage{caption}
\usepackage{cuted}
\usepackage{pifont}
\usepackage{fontawesome}
\newcommand{\xmark}{\ding{55}}

\usepackage[square,numbers]{natbib}

\usepackage[usenames,dvipsnames]{xcolor}
\usepackage{tcolorbox}
\definecolor{linkblue}{HTML}{2A1B81}
\hypersetup{
  colorlinks,
  citecolor=green,
  linkcolor=linkblue,
  urlcolor=linkblue}

\title{Continual learning of longitudinal health records}

\date{October 20, 2021}	

\author{ \href{https://orcid.org/0000-0003-4349-4453}{\includegraphics[scale=0.06]{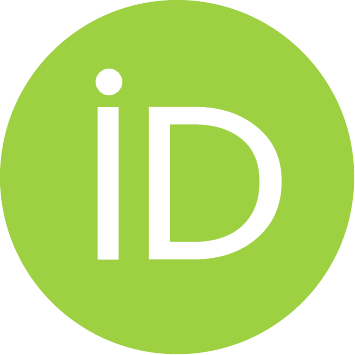}\hspace{1mm}Jacob Armstrong}\\
    Institute of Biomedical Engineering\\
	Oxford University\\\\
	\texttt{jacob.armstrong@eng.ox.ac.uk} \\
	
	\And
	 
	David A. Clifton \\
    Institute of Biomedical Engineering\\
	Oxford University\\\\
	\texttt{davidc@robots.ox.ac.uk} \\
}

\renewcommand{\headeright}{Preprint}
\renewcommand{\undertitle}{Preprint}
\renewcommand{\shorttitle}{Continual learning of longitudinal health records}

\hypersetup{
pdftitle={Continual learning of longitudinal health records},
pdfsubject={cs.LG, q-bio.QM},
pdfauthor={Jacob Armstrong},
pdfkeywords={Continual learning, EHR, Longitudinal data, Panel data, RNN, LSTM, MIMIC},
}

\begin{document}

\begin{strip}

\maketitle

\begin{abstract}


	
	\textit{Continual learning} denotes machine learning methods which can adapt to new environments while retaining and reusing knowledge gained from past experiences. Such methods address two issues encountered by models in non-stationary environments: ungeneralisability to new data, and the catastrophic forgetting of previous knowledge when retrained. This is a pervasive problem in clinical settings where patient data exhibits covariate shift not only between populations, but also continuously over time. However, while continual learning methods have seen nascent success in the imaging domain, they have been little applied to the multi-variate sequential data characteristic of critical care patient recordings.	Here we evaluate a variety of continual learning methods on longitudinal ICU data in a series of representative healthcare scenarios. We find that while several methods mitigate short-term forgetting, domain shift remains a challenging problem over large series of tasks, with only replay based methods achieving stable long-term performance. \\
	
	Code for reproducing all experiments can be found at \url{https://github.com/iacobo/continual}
	
\end{abstract}

\bigskip

\bigskip 
\bigskip
\end{strip}

\section{Introduction}


Clinical and healthcare-related machine learning studies have grown rapidly in recent years, with over a thousand publications annually since 2018 \citep{weissler2021role}. However many models suffer from ungeneralisability: the distribution of their training data is not representative of the setting in which they are deployed, and hence their real-world performance and utility is overestimated. Further, the distribution of data in a given environment itself continually shifts with time, limiting the use even of models trained on initially representative domains \citep{kelly2019key, futoma2020myth}.

\begin{figure}[t]
	\centering
	\includegraphics[width=\columnwidth]{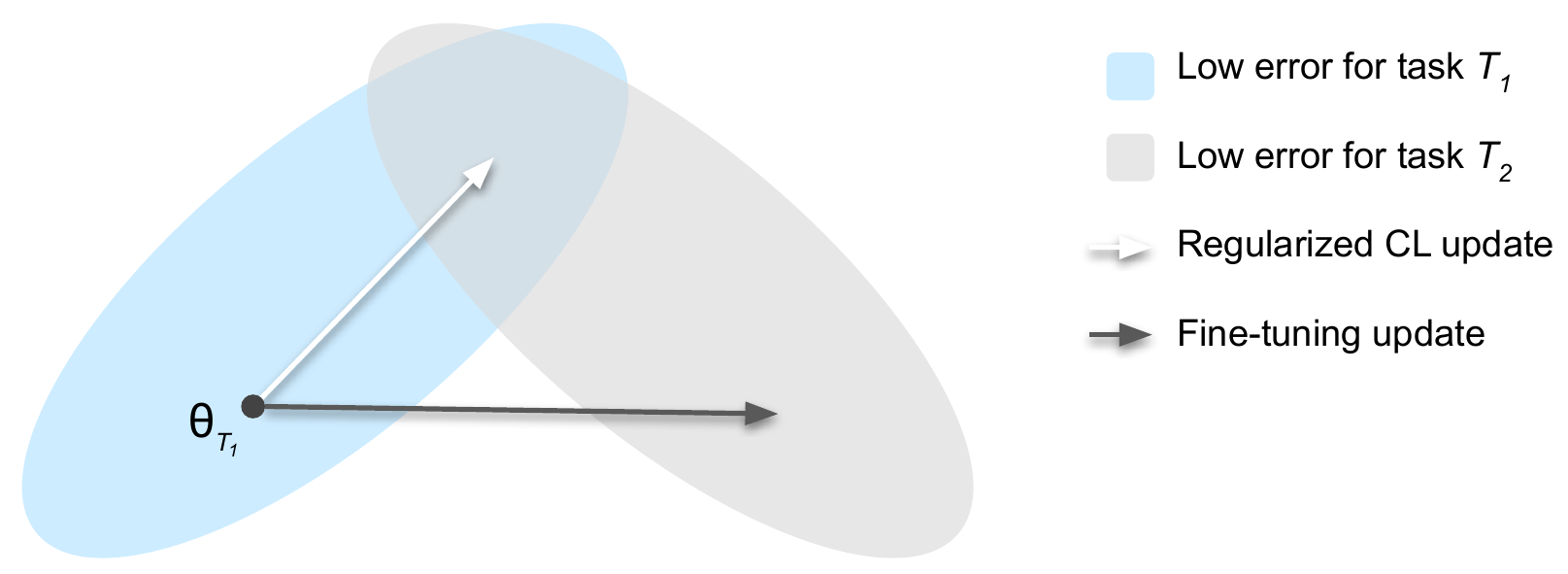}
	\caption{Under naive transfer learning (grey arrow), there is no guarantee that the parameter values ($\theta_{T_1}$) remain within a region of low error for the previous task $T_1$ (blue oval) after training on subsequent task $T_2$. Regularization techniques like Elastic Weight Consolidation (EWC) enforce such behaviour by penalising the loss, constraining parameter updates to a locus of learned values for previous tasks (figure adapted from \cite{kirkpatrick2017overcoming}).}
	\label{fig:param-update}
\end{figure}

Unfortunately, naively retraining networks on new data as it becomes available ("fine tuning") commonly results in forgetting of past knowledge. Models can overfit to the specific features of the new dataset, degrading performance on previous tasks in a process known as \textit{catastrophic forgetting}. This occurs since training on the current task propels updated parameter values far from the previously optimized values (see fig \ref{fig:param-update}). This effectively overwrites learned features pertinent to previous tasks when they are not useful for the current one. While accumulating data and periodically retraining models theoretically alleviates catastrophic forgetting, such approaches are practically encumbered by privacy, storage, and computational hurdles.

Continual learning (CL) has recently emerged as a field to tackle these issues. Models are designed to incrementally update on new datasets while retaining and reusing past knowledge where relevant. Concretely this refers to models which can sequentially train on a series of tasks, while retaining predictive power on previously encountered examples. 

However a number of state of the art techniques rely on storing past examples and hence may be infeasible in clinical settings due to privacy or data storage limitations. Generative models which create simulated pseudo-examples face further issues of computational limitations.


Further, while a large proportion of Electronic Health Records (EHR) used in patient monitoring and prognostics consists of periodic tabular readings (i.e. multi-variate time-series), most current evaluations of continual learning methods are in the image domain \citep{lee2020clinical, baweja2018towards}. Current benchmarks do not adequately capture the realistic issues faced in a clinical context (e.g. highly imbalanced classes, large multivariate sequences, sparse recordings) \citep{delange2021continual}, and hence the generalisability of their results to these contexts is unclear.

\paragraph{Contributions} In this work we present a set of representative continual learning scenarios in the medical domain derived from the open-access eICU-CRD and MIMIC-III ICU datasets \citep{johnson2016mimic, eicu-crd-2.0, goldberger2000physiobank}. We evaluate a range of methods on these problems, the first (to our knowledge) comprehensive study of Continual Learning methods on medical time-series data. Benchmarks demonstrate common domain shifts encountered by clinical systems in the real world, across geographies, time, and population demographics.

\paragraph{Related work} \citet{cossu2021continual} present a comprehensive evaluation of methods on a set of proposed benchmarks for sequence data. We extend on this work by evaluating such methods on real-world clinical scenarios, over a broader array of model architectures (including Transformers and alternative recurrent networks). \citet{aljundi2019task} examine imbalanced classification problems and the effect of dropout regularisation but from a task incremental perspective on imaging data. \citet{kiyasseh2020continual} investigate domain incremental learning on univariate physiological signals but examine only replay based methods. \citet{churamani2021domain} investigate domain incremental learning across ethnicity and gender but for facial image data, only evaluating regularization based methods. \citet{guo2021evaluation} and \citet{alves2018dynamic} investigate temporal and institutional domain shift in ICU data, but from a domain adaptation perspective, considering only a single source and target dataset.


\section{Background}

\subsection{Continual Learning Scenarios}

The typical continual learning problem consists of a model encountering a sequence of discrete batches of data, corresponding to different `tasks', where data cannot be stored between tasks.\footnote{More general settings exists in which models encounter a stream of incoming data, sometimes referred to as \textit{online learning}. However, since many CL methods are not designed for such settings we stick to the batch case to allow a broader comparison of methods.} For example a clinical decision model updated annually on new hospital data. The data cannot be retained longer than this due to privacy limitations, but we aspire for the model to generalise to the population with each dataset encountered, and not overfit to the most recent batch as is seen in traditional supervised learning.

Problems are typically split into three scenarios \citep{van2019three}:

\begin{itemize}
  \item \textbf{Task Incremental} Here each task is nominally different. In a classification setting this typically corresponds to each pair of tasks having non-overlapping target sets $Y_i \cap Y_j = \emptyset$ $\forall i \neq j$.
  \item \textbf{Class Incremental} Here the set of potential targets expands with each task: $Y_i \subset Y_j$ $\forall i<j$.
  \item \textbf{Domain Incremental} Here tasks are nominally the same (i.e. the set of targets is identical for all tasks $Y_i = Y_j$ $\forall i \neq j$), but the distribution of input-features changes with each task.
\end{itemize}

However, as noted by \citet{cossu2021continual}, this does not capture the full breadth of potential scenarios. For example, newly encountered datasets may introduce a mix of domain shifted instances of old classes, new classes, or novel combinations of classes. \citet{maltoni2019continuous} divide scenarios into multi-task, single-incremental task, and multiple-incremental tasks, along with a secondary classification for new examples containing new instances of old classes, new classes, or both. However, as discussed by \cite{cossu2021continual}, several of the proposed categories are unrealistic or rare. For simplicity we use the terminology of \citet{van2019three}.



\subsection{Ontology of methods}

A number of methods have been proposed in recent years to mitigate catastrophic forgetting, falling under three general archetypes \citep{parisi2019continual}:

\begin{itemize}
  \item \textbf{Regularization} A regularization constraint is added to the loss function, enforcing updated parameter values to lie within a radius of the current value. This has the benefit of a natural Bayesian interpretation where the posterior values after training on task $T_i$ inform the priors for task $T_{i+1}$. Methods differ in strategies for choosing which parameters to constrain, and to what degree.
  \item \textbf{Rehearsal} A subset of examples (or generated pseudo-examples) from previous tasks are cached and mixed in with each new task's training set. Methods differ chiefly in the criteria used for choosing examples. Also known as \textit{replay}.
  \item \textbf{Dynamic architectures} A broad variety of techniques where the network architecture itself adapts with new task presentation. Approaches range from hyper-networks with task-specific subnetworks, to initially small networks which add neurons as resources are required to model new tasks. They are broadly characterised by increasing network complexity with number of tasks.
\end{itemize}

Such architectural features are not mutually exclusive, and may be hybridised in a number of ways. For example, GEM \citep{lopez2017gradient}, iCARL \citep{rebuffi2017icarl}, and FRoMP \citep{pan2020continual} employ both rehearsal and regularisation elements. More complex ontologies have been proposed to finer categorise such methods \citep{delange2021continual}.

Replay methods achieve state of the art in many scenarios examined in the literature \citep{riemer2018learning, chaudhry2019tiny}. However, such techniques are often infeasible in real-world settings, where previous examples cannot be stored or shared due to data privacy constraints \citep{lee2020clinical, farquhar2018towards}. Such a problem is not unique to clinical settings, and while \textit{generative} replay models simulating past examples have been proposed \citep{shin2017continual}, sparse and complex sequential data can prohibit learning of an adequate generative distribution function \citep{ehret2020continual}.

For an in depth review of continual learning methods generally, we refer to \citet{delange2021continual, parisi2019continual, luo2020appraisal}. For convenience, we briefly outline the methods evaluated in this work below:

\paragraph{Regularization approaches}
\begin{itemize}
    \item \textbf{Elastic Weight Consolidation (EWC)} \citep{kirkpatrick2017overcoming} Penalises changes in parameter values relative to the \textit{importance} of parameters to previous task(s). Importances determined via Fisher's information matrix. Parameters which are important to previous task(s) are highly constrained, and ones of less importance are less constrained during updates.
    
    \item \textbf{Online EWC} \citep{schwarz2018progress} An adaptation of EWC using a running average of task importance penalties, as opposed to distinct penalties for each previous task. Computationally more efficient and tractable for a large number of tasks.
    
    
    \item \textbf{Synaptic Intelligence (SI)} \citep{zenke2017continual} Similar to EWC, enforces parameter specific regularization but importances are calculated \textit{online} (i.e. during training) by approximating the effect on loss and gradient update, as opposed to during an additional pass of the network post training.
    
    \item \textbf{Learning without Forgetting (LwF)} \citep{li2017learning} A copy of the model parameters before updating on the current task is stored and compared to the updated version. Parameter values are distilled between both versions for final update. Hence may be categorised as a \textit{functional} regularization strategy.
    
\end{itemize}

\paragraph{Replay approaches}

\begin{itemize}
    \item \textbf{Replay} Naive storage of a set of random examples per task, which are mixed in with each subsequent task's training data. May employ more specific storage policies such as class or task-wise balancing of memories.
    
    \item \textbf{GDumb} \citep{prabhu2020gdumb} A greedy rehearsal method in which the memory buffer is filled with the $\frac{\textrm{buffer size}}{\textrm{$n$ tasks seen}}$ most recently encountered examples per task. Examples replayed with each new task.

    \item \textbf{Gradient Episodic Memory (GEM)} \citep{lopez2017gradient} Stores a set of examples from each task. Selectively updates gradient for a given minibatch on the current task only if the gradient can be projected in a plane which maintains the positivity of the gradient updates for all stored examples.
    
    \item \textbf{Averaged Gradient Episodic Memory (A-GEM)} \citep{chaudhry2018efficient} Adaptation of GEM considering only the average gradient for a randomly sampled subset of the stored examples.
\end{itemize}

\paragraph{Dynamic approaches}

\begin{itemize}
    \item \textbf{Progressive Neural Network (PNN)} \citep{rusu2016progressive} A copy of parameter weights before updating on a new task is stored. If any parameters shift beyond a certain threshold, the previous weights are frozen and cloned to produce a sister neuron with the updated weights. Relies on task identity at inference to ensure shifted `sister' neurons do not interfere with predictions for prior tasks.
\end{itemize}


\section{Experiments}
\label{sec:experiments}

\subsection{Problem definitions}
\paragraph{Domain Incremental}

We consider 3 natural Domain Incremental experiments, corresponding to $n$ patient ICU datasets encountered sequentially across time or location. Domain increments correspond to changing:

\begin{itemize}
    \item time (season) $(n=4)$
    \item hospital $(n=155)$
    \item region $(n=4)$
\end{itemize}

We also consider the following 3 artificial Domain Incremental experiments, simulating imbalanced populations between healthcare environments (due to demographic-specific care in a given institution, or general population imbalance). Domain increments correspond to groups of patients split by:

\begin{itemize}
    \item age group $(n=7)$
    \item ethnicity (broad) $(n=5)$
    \item ICU ward $(n=8)$
\end{itemize}

The majority of the above domain splits are self explanatory. \textsc{ICU ward} refers to different types of critical care (i.e. intensive care) unit, which may specialise in cardiac, trauma, neurological etc injuries.


For each task the setting is supervised prediction of a binary outcome (48hr in-hospital mortality). Input data are multivariate time-series, consisting of periodically recorded patient vital signs from an ICU admission. These are sampled at a rate of 1 per hour, and are of duration $t = 48$ time steps. Static covariates are repeated to the length of the time-varying sequence and concatenated to enable processing by sequential models.


Further experiments on alternative outcomes (acute respiratory failure; shock) and different sequence/prediction window lengths ($t \in \{4,12\})$ can be found in Appendix \ref{sec:full-results}.

Note that \textsc{region} and \textsc{ethnicity (broad)} can be seen as easier, lower resolution versions of the \textsc{hospital} and \textsc{ethnicity (narrow)} experiments respectively since their domains correspond to non-overlapping supersets of the formers'.






\begin{tcolorbox}[colback=red!5,colframe=red!75!black,title=Work in progress \faWrench]

\textbf{Note:} not all experiments described have been completed (e.g. PNN strategy, Region experiment, transformer architecture for \textsc{hospital}, class incremental experiments, class incremental experiment, and supplementary experiments on memory size, traditional regularization, alternative outcome definitions, and sequence length). 

\end{tcolorbox}

\subsection{Experimental setup}

\begin{figure}[b]
	\centering
    	\includegraphics[width=\columnwidth]{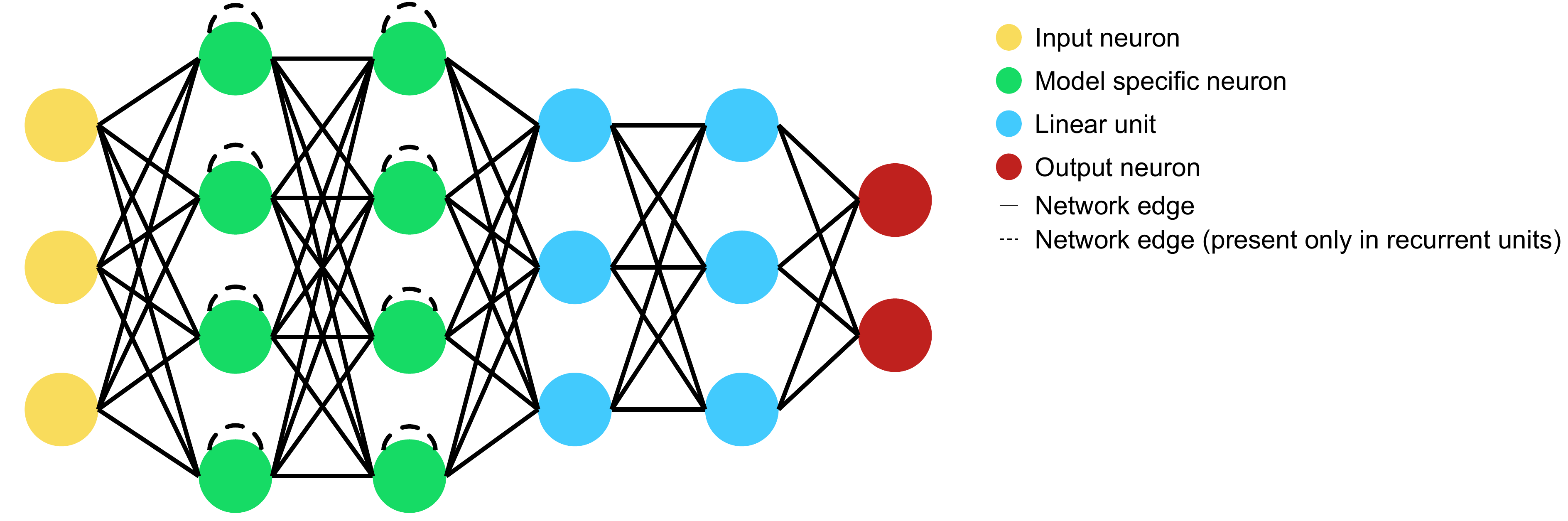}
    	\caption{Generic structure of binary classification model. Example contains 2 hidden `feature' layers of width 4 (green), and 2 fully connected `classification' layers of width 3 (blue).}
    	\label{fig:models}
\end{figure}

\paragraph{Model architectures} For each problem, we evaluate 4 basic neural network architectures: 

\begin{enumerate}
    \item a dense feedforward network (MLP)
    \item 1d convolutional neural network (CNN)
    \item long- short-term memory network (LSTM); and
    \item transformer 
\end{enumerate}

These were chosen to give a breadth of sequential models, along with a data-structure agnostic model (MLP) for baseline comparison. Further recurrent models are evaluated in Appendix \ref{sec:additional-sequential-models}. Models consist of 1 to 4 architecture-specific layers, followed by two dense linear layers (see fig \ref{fig:models}). 

To enable more fair comparison of methods, model-level parameters (such as number and width of layers) were tuned for the naive baselines and frozen for all other methods (for a given architecture and experiment). Standard regularization features such as dropout were omitted to clearer investigate the effect of the continual learning mechanisms themselves. Additional experiments on the effect of regularization strategies on catastrophic forgetting are found in Appendix \ref{sec:dropout}. Batch Normalisation was not used in the CNN due to its intensifying effect on catastrophic forgetting \citep{lomonaco2020rehearsal}.

\paragraph{Strategies} Each model is equipped with one of the 8 continual learning strategies listed in Table \ref{tab:methods}:

\begin{table}[ht]
\centering
\begin{adjustbox}{max width=\columnwidth}
\begin{tabular}{llll}
\textbf{Archetype} & \textbf{Method}                   & \textbf{Abbreviation} & \textbf{Source} \\ 
\hline
\multirow{2}{*}{Baseline} 
                   & Naive fine-tuning                 & Naive       &                 \\
                   & Cumulative multi-task training    & Cumulative  &                 \\ \\
\multirow{3}{*}{Regularization} 
                   & Elastic Weight Consolidation      & EWC         & \citep{kirkpatrick2017overcoming}     \\
                   & Online EWC                        & Online EWC  & \citep{schwarz2018progress}     \\
                   & Synaptic Intelligence             & SI          &   \citep{zenke2017continual}              \\
                   & Learning without Forgetting       & LwF         &   \citep{li2017learning}              \\ \\
\multirow{4}{*}{Rehearsal}
                   & Naive replay                      & Replay      &                 \\
                   & GDumb                             & GDumb       & \citep{prabhu2020gdumb}          \\
                   & Gradient Episodic Memory          & GEM         & \citep{lopez2017gradient}    \\
                   & Averaged Gradient Episodic Memory & AGEM        & \citep{chaudhry2018efficient}             
\end{tabular}
\end{adjustbox}
\caption{Continual Learning methods evaluated.}
\label{tab:methods}
\end{table}

Rehearsal based methods are given a fixed budget of 256 samples per task, corresponding to approximately 5\% and 0.5\% of the training data for MIMIC and eICU experiments respectively. See Appendix \ref{sec:buffer-size} for experiments on increasing storage capacity.

We further evaluate all models using two baseline methods:

\begin{itemize}
    \item \textbf{Naive:} Naive fine-tuning on each additional task. This is a soft lower bound on performance, equivalent to serial transfer learning with no continual learning mechanism. It is expected to undergo catastrophic forgetting.
    \item \textbf{Cumulative:} Cumulative multi-task training on all tasks seen thus far. This is a soft upper bound on performance, equivalent to transfer learning on a continually expanding dataset, or a rehearsal method with unlimited storage capacity. Note that continual learning methods may outperform this in the instance of strong backwards transfer of information, or on tasks with considerable imbalance in dataset sizes.
\end{itemize}


\paragraph{Data} We use the open-access \textbf{eICU-CRD} \cite{eicu-crd-2.0} ICU dataset for all experiments bar seasonal and narrow ethnicity domain increments, for which such information was not available. For these we use the open-access \textbf{MIMIC-III} \cite{johnson2016mimic, goldberger2000physiobank} ICU database. For standardisation of preprocessing and outcome definitions, datasets were preprocessed with the \textbf{FIDDLE} pipeline \citep{mimic21shengpu}. Data can be accessed at \url{https://www.physionet.org/content/mimic-eicu-fiddle-feature/1.0.0/}.

Relevant domain shifts identifiable in both datasets is listed in Table \ref{tab:datasets}. Full list of domain shifts, along with number of samples in each task, domain, and train/validation/test partition are available in Appendix Table \ref{tab:partition}.

\begin{table}[ht]
\begin{adjustbox}{max width=\columnwidth}
\begin{tabular}{cclr}
\textbf{MIMIC-III} & \textbf{eICU} & \textbf{Domain increment}   & \textbf{Number of domains} \\
\hline
                   & \checkmark    & Region (US)                 & 4    \\
                   & \checkmark    & Hospital                    & 155  \\
\checkmark         & \checkmark    & Unit                        & 5-8    \\
\checkmark         & \checkmark    & Sex                         & 2    \\
\checkmark         & \checkmark    & Age                         & 6-7  \\
\checkmark         & \checkmark    & Ethnicity (broad)           & 5    \\
\checkmark         &               & Ethnicity (narrow)          & 20   \\
\checkmark         &               & Time (season)               & 4    \\
\end{tabular}
\end{adjustbox}
\caption{Domain shifts annotated in the MIMIC-III and eICU-CRD datasets. When a range of values are given, these correspond to different domains represented across sub-populations with different outcomes (i.e. mortality, Shock, ARF) or datasets (MIMIC-III, eICU).}
\label{tab:datasets}
\end{table}

\paragraph{Metrics} We compared the methods using Balanced Accuracy as the main metric.


Since class sizes are highly imbalanced in all experiments (mortality outcome averaging 10\% across tasks, see Table \ref{tab:partition}), and the degree of class imbalance is not constant across domain splits, accuracy is an inappropriate measure of model performance \citep{roy2021multi}. In minority-event detection, metrics such as sensitivity and specificity (i.e. true positive and true negative rates) are often preferred depending on the relative importance of Type I and Type II errors in the given medical context \citep{hicks2021evaluation}. To simplify presentation of results, we report the Balanced Accuracy, an average of specificity and sensitivity. Full presentation of sensitivity, specificity, precision, class-accuracy, Area Under the Receiver Operating Curve (AUROC), and Area Under the Precision Recall Curve (AUPRC) can be found in Appendix \ref{sec:full-results}.

\subsubsection{Pipeline}

\begin{enumerate}
    \item \textbf{Task split} Data is initially split into several tasks via the task identity (i.e. demographic category for Domain Incremental experiments). For some demographics there were no positive outcomes (e.g. some low volume ethnicity groups). These groups were excluded from the dataset for that experiment. Task order was randomized.
    
    \item \textbf{Train, validation, test split} Data within each task is then split into train, validation, and test subsets for the first two tasks, and into train, test subsets only for all subsequent tasks in proportions 70:15:15 and 70:30 respectively. Since multiple ICU admissions can pertain to the same patient, train/validation/test streams were split along patient identities to avoid data leakage of similar records \citep{kaufman2012leakage}. Sample counts for each experiment can be found in Table \ref{tab:partition}.
    
    \begin{figure}[t!]
        \centering
        \includegraphics[width=\columnwidth]{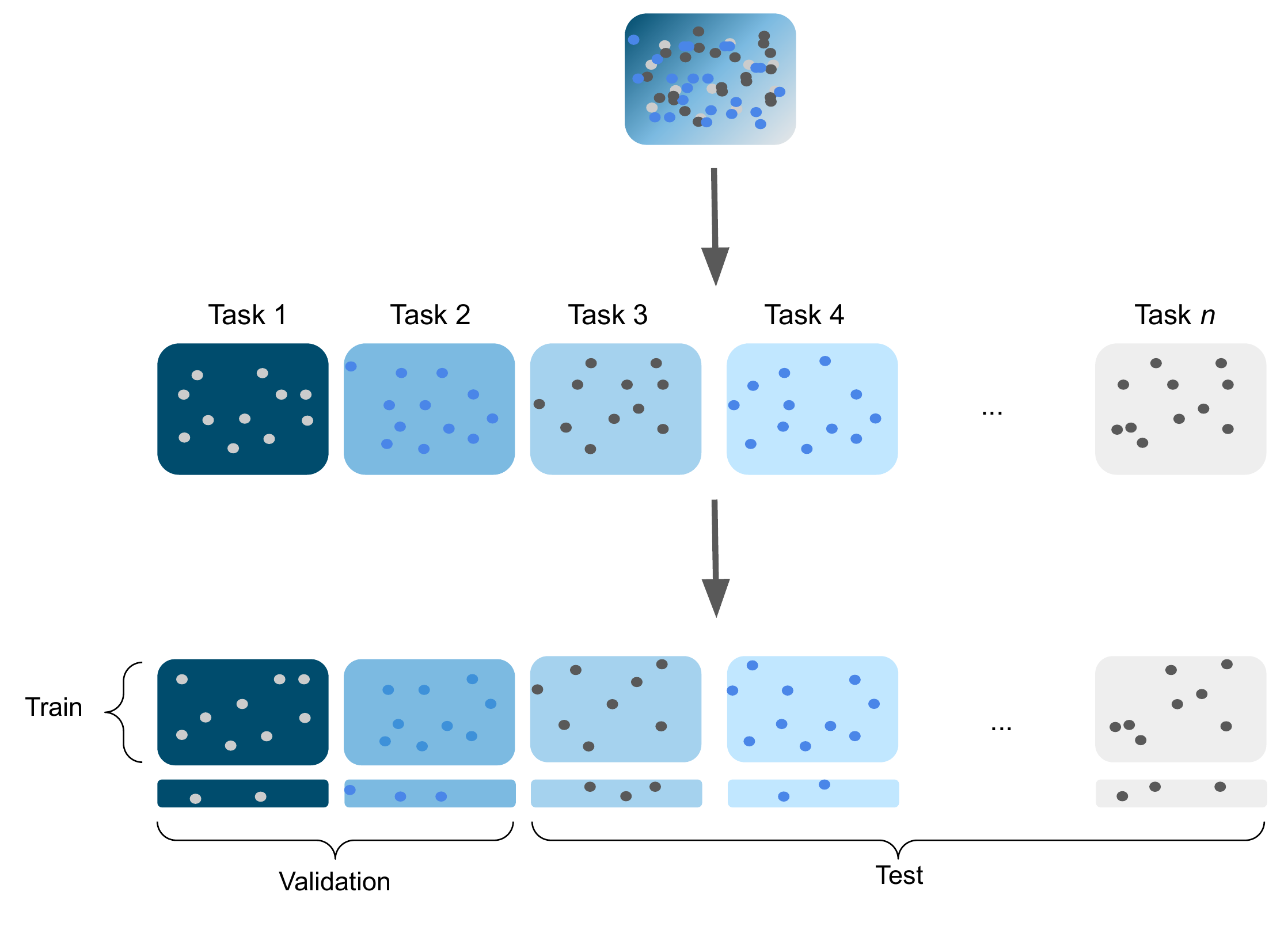}
    	\caption{Data is initially divided into sequential `tasks' split by domain shift. Task order is randomized. The first two tasks are split into training (85\%) and validation (15\%) sets, the latter used for hyperparameter tuning. Subsequent tasks are assumed to be unavailable for hyperparameter tuning and are split into training (85\%) and test data (15\%) only. Different colours refer to different domain shifts within the complete dataset.}
    	\label{fig:figs/pipeline}
    \end{figure}
    
    \item \textbf{Hyperparameter optimisation} Hyperparameter optimisation requires careful consideration in a continual learning setting, since we should not have access to validation sets from future tasks during the hypothesis generation phase (i.e. model specification). As such, tuning was performed using validation data from the first two tasks only. For setups with a large ($>5 $) number of tasks, these first two tasks are excluded from the final training and testing phase. Otherwise, training and validation data are combined at this stage. This setup is consistent with validation regimes proposed in \citep{chaudhry2018efficient} for Continual Learning setups with a limited number of tasks.

    For fairer comparison of methods, generic hyperparameters (i.e. learning rate, batch size, number of layers, hidden depth) were tuned for the Naive baseline run only and frozen for all other methods. Strategy specific hyperparameters were tuned independently for each method.

    Hyperparameters were sampled from a range of reasonable values determined from the literature \citep{mimic21shengpu, cossu2021continual}. Where methods shared identical or analogous parameters, the search-space was also shared to ensure fair comparison (for example, regularization strength in EWC, SI, and LwF). A full list of hyperparameter search spaces and the best performing configurations for each model can be found in Appendix \ref{sec:hyperparameters}.
    
    Hyperparameters were chosen which maximised the average balanced accuracy of the validation predictions for the first two tasks.
    
    \begin{figure*}[hb!]
\begin{adjustbox}{max width=\textwidth}
	\centering
    	\includegraphics[width=\textwidth]{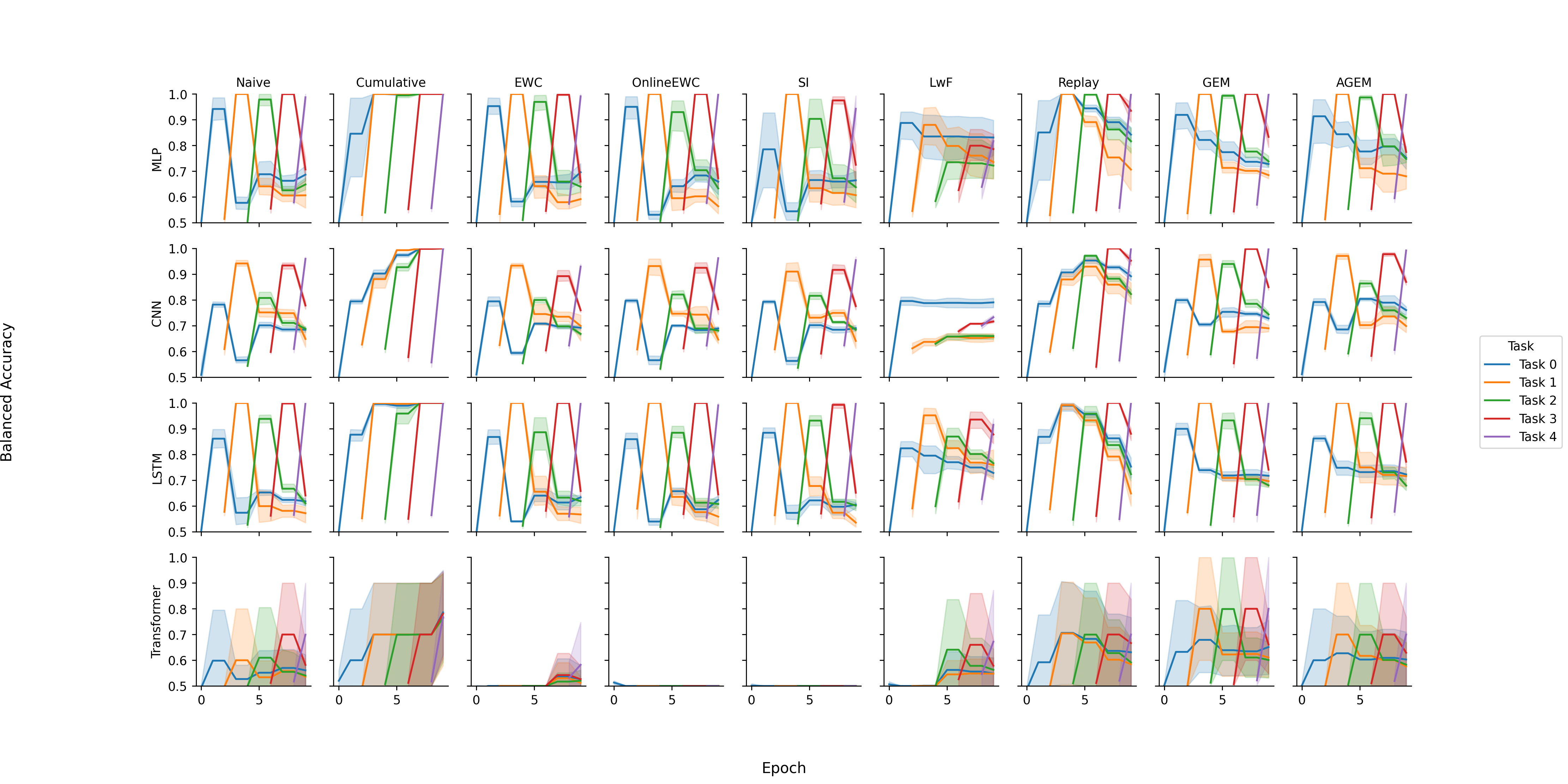}
\end{adjustbox}
\caption{Domain Incremental results for an outcome of mortality (48h) across domain shift of different \textsc{ICU Ward}. Coloured lines show the training balanced accuracy for each task as it is encountered, for each model and strategy. Shaded regions (left) and black bars (right) refer to bootstrapped 95\% confidence intervals. Naive methods (first column) notably undergo catastrophic forgetting as new tasks are introduced. Cumulative training (second column) mitigates this. Regularization methods mitigate this to a degree for the most recently encountered task(s), but do not maintain performance across the entire history of tasks (with the notable exception of LwF). Regularization techniques appear most effective in combination with CNN's. Rehearsal methods achieve greater success across a range of architectures, achieving top performance in most experiments.}
\label{fig:exp-results}
\end{figure*}

    \item \textbf{Training} Once hyper-parameters were selected, each model/strategy combination was trained from scratch on the sequence of tasks' training data. In Appendix \ref{sec:curriculum} we present an extra experiment on the impact of task ordering (cf. \textit{curriculum learning}). The objective function of training was minimising the weighted cross entropy of predictions. Weights are determined by the inverse proportion of class examples in the first two tasks' training data.
    
    \item \textbf{Evaluation} Models were evaluated on each task's test data, with balanced accuracy, forgetting, and weighted cross entropy loss recorded. Per-task and average metrics were recorded at the end of each training epoch. Training and evaluation was repeated from random initialisation 5 times. Mean performance and bootstrapped 95\% confidence intervals are reported.
\end{enumerate}


\clearpage

\section{Results}
\label{sec:results}




\begin{table*}[t!]

\begin{adjustbox}{max width=\textwidth}
\begin{tabular}{llllll|llll|llll|llll}
{} & {} & \multicolumn{4}{c}{\textbf{\textsc{Age}}} & \multicolumn{4}{c}{\textbf{\textsc{Ethnicity (broad)}}} & \multicolumn{4}{c}{\textbf{\textsc{ICU Ward}}} & \multicolumn{4}{c}{\textbf{\textsc{Time (season)}}} \\
{} & {} & {CNN} & {LSTM} & {MLP} & {Transformer} & {CNN} & {LSTM} & {MLP} & {Transformer} & {CNN} & {LSTM} & {MLP} & {Transformer} & {CNN} & {LSTM} & {MLP} & {Transformer} \\
\midrule
\multirow[c]{2}{*}{Baseline} & Cumulative & 64.9$_{\pm1.6}$ & 64.3$_{\pm0.8}$ & 63.0$_{\pm6.4}$ & 59.0$_{\pm4.9}$ & 61.6$_{\pm1.1}$ & 60.6$_{\pm1.2}$ & 60.6$_{\pm0.8}$ & 53.7$_{\pm4.8}$ & 59.2$_{\pm0.9}$ & 58.2$_{\pm1.8}$ & 58.8$_{\pm1.0}$ & 57.3$_{\pm4.1}$ & 64.1$_{\pm1.1}$ & 65.9$_{\pm1.4}$ & 65.5$_{\pm2.0}$ & 53.5$_{\pm6.8}$ \\
 & Naive & 64.0$_{\pm0.7}$ & 62.8$_{\pm1.2}$ & 50.0$_{\pm0.0}$ & 57.8$_{\pm3.8}$ & 67.6$_{\pm0.9}$ & 67.6$_{\pm0.7}$ & 68.8$_{\pm0.9}$ & 50.0$_{\pm0.0}$ & 64.2$_{\pm1.7}$ & 56.6$_{\pm2.0}$ & 58.7$_{\pm1.2}$ & 53.7$_{\pm4.4}$ & 67.2$_{\pm2.3}$ & 67.8$_{\pm1.5}$ & 67.6$_{\pm1.0}$ & 50.0$_{\pm0.0}$ \\
 
& & & & & & & & & & & & & & & & & \\ 
 
\multirow[c]{4}{*}{Regularization} & EWC & 63.8$_{\pm1.3}$ & 63.6$_{\pm0.9}$ & 50.0$_{\pm0.0}$ & 55.7$_{\pm5.7}$ & 67.3$_{\pm1.1}$ & 66.8$_{\pm1.6}$ & 69.1$_{\pm0.4}$ & 53.5$_{\pm6.8}$ & 62.9$_{\pm1.1}$ & 58.8$_{\pm3.5}$ & 58.6$_{\pm1.3}$ & 51.9$_{\pm3.7}$ & 66.4$_{\pm1.2}$ & 66.9$_{\pm0.9}$ & 68.0$_{\pm0.6}$ & 50.0$_{\pm0.0}$ \\
 & LwF & \bfseries 64.7$_{\pm0.9}$ & \bfseries 64.4$_{\pm0.8}$ & 64.3$_{\pm0.7}$ & 58.4$_{\pm4.2}$ & 67.5$_{\pm0.2}$ & \bfseries 67.8$_{\pm1.1}$ & 69.2$_{\pm0.6}$ & 52.6$_{\pm5.0}$ & 64.3$_{\pm0.5}$ & \bfseries 61.8$_{\pm1.0}$ & \bfseries 62.4$_{\pm2.3}$ & \bfseries 54.5$_{\pm5.5}$ & 67.1$_{\pm2.0}$ & 67.0$_{\pm1.4}$ & 67.8$_{\pm0.8}$ & 52.4$_{\pm4.7}$ \\
 & OnlineEWC & 63.7$_{\pm0.8}$ & 63.1$_{\pm0.6}$ & \bfseries 64.6$_{\pm0.5}$ & \bfseries 61.1$_{\pm0.7}$ & \bfseries 67.8$_{\pm0.5}$ & 66.7$_{\pm1.7}$ & \bfseries 70.0$_{\pm0.8}$ & 53.2$_{\pm6.3}$ & 64.2$_{\pm0.8}$ & 59.5$_{\pm2.4}$ & 58.9$_{\pm1.9}$ & 50.0$_{\pm0.0}$ & 67.7$_{\pm1.1}$ & \bfseries 67.8$_{\pm0.5}$ & 68.1$_{\pm0.4}$ & 50.0$_{\pm0.0}$ \\
 & SI & 63.9$_{\pm1.4}$ & 62.8$_{\pm1.9}$ & 63.7$_{\pm0.3}$ & 57.3$_{\pm4.3}$ & 67.5$_{\pm1.0}$ & 67.3$_{\pm1.6}$ & 69.9$_{\pm0.4}$ & 54.8$_{\pm6.5}$ & \bfseries 64.5$_{\pm0.5}$ & 58.9$_{\pm1.1}$ & 60.6$_{\pm1.8}$ & 50.0$_{\pm0.0}$ & 66.1$_{\pm0.6}$ & 67.6$_{\pm0.6}$ & 67.6$_{\pm0.6}$ & 52.7$_{\pm5.4}$ \\
 
& & & & & & & & & & & & & & & & & \\ 
 
\multirow[c]{3}{*}{Rehearsal} & AGEM & 64.5$_{\pm1.0}$ & 62.2$_{\pm0.9}$ & 64.1$_{\pm0.6}$ & 58.0$_{\pm4.1}$ & 64.8$_{\pm2.2}$ & 67.3$_{\pm1.5}$ & 68.7$_{\pm0.2}$ & \bfseries 56.1$_{\pm7.3}$ & 63.9$_{\pm1.3}$ & 59.2$_{\pm1.5}$ & 60.8$_{\pm0.9}$ & 53.9$_{\pm4.7}$ & \bfseries 68.4$_{\pm1.6}$ & 67.2$_{\pm2.1}$ & \bfseries 68.6$_{\pm0.9}$ & 50.0$_{\pm0.0}$ \\
 & GEM & 63.1$_{\pm0.8}$ & 60.6$_{\pm1.1}$ & 61.7$_{\pm0.6}$ & 58.5$_{\pm1.4}$ & 58.2$_{\pm1.1}$ & 57.8$_{\pm1.1}$ & 60.2$_{\pm0.4}$ & 50.8$_{\pm1.6}$ & 60.3$_{\pm1.6}$ & 57.4$_{\pm1.5}$ & 57.3$_{\pm1.3}$ & 53.8$_{\pm3.2}$ & 60.1$_{\pm1.1}$ & 60.1$_{\pm2.4}$ & 63.7$_{\pm0.9}$ & 54.4$_{\pm5.2}$ \\
 & Replay & 60.0$_{\pm1.2}$ & 58.1$_{\pm1.8}$ & 51.1$_{\pm2.2}$ & 59.0$_{\pm1.6}$ & 61.6$_{\pm3.7}$ & 60.3$_{\pm3.6}$ & 61.6$_{\pm2.1}$ & 51.5$_{\pm3.0}$ & 59.0$_{\pm1.7}$ & 55.7$_{\pm1.6}$ & 58.7$_{\pm1.5}$ & 53.2$_{\pm3.8}$ & 65.9$_{\pm3.0}$ & 61.4$_{\pm2.3}$ & 65.2$_{\pm1.8}$ & \bfseries 55.6$_{\pm4.7}$ \\
 
\addlinespace[1cm]
 
\end{tabular}
\end{adjustbox}

\begin{adjustbox}{max width=\textwidth}
\begin{tabular}{lllll|lll|lll|lll|lll}
{} & {} & \multicolumn{3}{c}{\textbf{\textsc{Hospital (7)}}} & \multicolumn{3}{c}{\textbf{\textsc{Hospital (14)}}} & \multicolumn{3}{c}{\textbf{\textsc{Hospital (21)}}} & \multicolumn{3}{c}{\textbf{\textsc{Hospital (28)}}} &
\multicolumn{3}{c}{\textbf{\textsc{Hospital (35)}}} \\
{} & {} & {CNN} & {LSTM} & {MLP} & {CNN} & {LSTM} & {MLP} & {CNN} & {LSTM} & {MLP} & {CNN} & {LSTM} & {MLP} & {CNN} & {LSTM} & {MLP} \\
\midrule
\multirow[c]{2}{*}{Baseline} & Cumulative & 57.3$_{\pm1.2}$ & 55.2$_{\pm0.8}$ & 56.5$_{\pm0.3}$ & 62.2$_{\pm2.5}$ & 61.6$_{\pm0.8}$ & 61.5$_{\pm0.3}$ & 57.9$_{\pm1.0}$ & 60.3$_{\pm1.2}$ & 60.9$_{\pm0.8}$ & 54.6$_{\pm0.6}$ & 55.5$_{\pm0.9}$ & 56.1$_{\pm0.7}$ & 56.0$_{\pm1.7}$ & 56.9$_{\pm1.5}$ & 56.1$_{\pm1.6}$ \\
 & Naive & 52.6$_{\pm0.1}$ & 52.4$_{\pm0.3}$ & 55.0$_{\pm0.1}$ & 57.4$_{\pm1.4}$ & 57.9$_{\pm1.8}$ & 61.9$_{\pm0.8}$ & 58.3$_{\pm1.8}$ & 57.0$_{\pm0.9}$ & 61.1$_{\pm0.9}$ & 52.0$_{\pm0.5}$ & 52.6$_{\pm0.7}$ & 54.1$_{\pm0.4}$ & 52.2$_{\pm0.4}$ & 52.0$_{\pm0.4}$ & 52.5$_{\pm0.1}$ \\
 
& & & & & & & & & & & & & & & & \\ 
 
\multirow[c]{4}{*}{Regularization} & EWC & 52.6$_{\pm0.0}$ & 52.5$_{\pm0.1}$ & 54.5$_{\pm1.1}$ & 57.9$_{\pm1.4}$ & 58.9$_{\pm0.5}$ & 61.2$_{\pm1.1}$ & 58.8$_{\pm1.7}$ & 57.4$_{\pm1.6}$ & 61.8$_{\pm1.0}$ & 52.4$_{\pm0.6}$ & \bfseries 54.7$_{\pm1.8}$ & 54.2$_{\pm0.4}$ & 51.9$_{\pm0.1}$ & 52.5$_{\pm0.8}$ & 52.5$_{\pm0.1}$ \\
 & LwF & 52.6$_{\pm0.1}$ & 52.6$_{\pm0.1}$ & 55.0$_{\pm0.1}$ & 56.6$_{\pm0.4}$ & 57.4$_{\pm1.0}$ & 61.1$_{\pm1.0}$ & 58.8$_{\pm1.2}$ & 57.8$_{\pm1.0}$ & 61.8$_{\pm0.9}$ & 51.8$_{\pm0.5}$ & 53.9$_{\pm0.9}$ & 54.1$_{\pm0.6}$ & 51.9$_{\pm0.1}$ & 51.8$_{\pm0.3}$ & 52.4$_{\pm0.0}$ \\
 & OnlineEWC & 52.6$_{\pm0.0}$ & 52.5$_{\pm0.1}$ & 55.0$_{\pm0.1}$ & 57.1$_{\pm0.7}$ & 58.6$_{\pm1.0}$ & 61.5$_{\pm1.1}$ & 58.1$_{\pm1.1}$ & 57.5$_{\pm1.9}$ & 61.1$_{\pm1.1}$ & 51.6$_{\pm0.5}$ & 53.6$_{\pm0.9}$ & 54.1$_{\pm0.5}$ & 52.2$_{\pm0.4}$ & 52.6$_{\pm0.9}$ & 52.6$_{\pm0.3}$ \\
 & SI & 52.6$_{\pm0.0}$ & \bfseries 53.7$_{\pm1.3}$ & 54.5$_{\pm1.0}$ & 58.3$_{\pm2.1}$ & 58.1$_{\pm1.1}$ & 61.6$_{\pm1.1}$ & 57.6$_{\pm0.7}$ & 57.9$_{\pm1.7}$ & 61.6$_{\pm0.7}$ & 51.7$_{\pm0.1}$ & 51.9$_{\pm0.7}$ & 53.6$_{\pm0.8}$ & 52.1$_{\pm0.4}$ & 52.4$_{\pm0.8}$ & 52.7$_{\pm0.2}$ \\
 
& & & & & & & & & & & & & & & & \\ 
 
\multirow[c]{3}{*}{Rehearsal} & AGEM & 52.3$_{\pm0.3}$ & 52.5$_{\pm0.1}$ & 56.1$_{\pm1.7}$ & 57.3$_{\pm1.7}$ & 57.6$_{\pm0.9}$ & \bfseries 62.9$_{\pm1.5}$ & \bfseries 59.5$_{\pm1.8}$ & 57.3$_{\pm3.4}$ & \bfseries 63.3$_{\pm0.6}$ & 51.9$_{\pm0.4}$ & 53.8$_{\pm1.5}$ & \bfseries 56.1$_{\pm0.9}$ & 52.2$_{\pm0.4}$ & 52.5$_{\pm0.8}$ & 52.9$_{\pm0.4}$ \\
 & GEM & \bfseries 54.8$_{\pm1.5}$ & 50.5$_{\pm1.0}$ & \bfseries 56.9$_{\pm1.3}$ & 58.2$_{\pm1.4}$ & 58.9$_{\pm1.6}$ & 61.0$_{\pm0.8}$ & 57.9$_{\pm1.8}$ & \bfseries 58.9$_{\pm0.9}$ & 59.3$_{\pm1.0}$ & \bfseries 53.0$_{\pm0.3}$ & 54.3$_{\pm0.6}$ & 55.4$_{\pm0.7}$ & \bfseries 54.0$_{\pm1.1}$ & \bfseries 55.5$_{\pm1.3}$ & \bfseries 58.1$_{\pm1.1}$ \\
 & Replay & 54.4$_{\pm1.3}$ & 53.2$_{\pm1.2}$ & 55.8$_{\pm1.9}$ & \bfseries 58.8$_{\pm1.4}$ & \bfseries 59.7$_{\pm0.4}$ & 62.1$_{\pm2.3}$ & 57.5$_{\pm1.2}$ & 56.9$_{\pm1.1}$ & 59.9$_{\pm1.8}$ & 52.5$_{\pm0.3}$ & 53.0$_{\pm0.6}$ & 53.8$_{\pm0.9}$ & 52.8$_{\pm1.0}$ & 52.9$_{\pm0.6}$ & 52.7$_{\pm0.1}$ \\
\end{tabular}
\end{adjustbox}
\caption{Final average balanced accuracy for 48hr mortality prediction across demographic domain shift (\textsc{Age, Ethnicity, Ward}) and \textsc{time} (top), and \textsc{hospital} shift (bottom). Average performance over 5 runs are presented with bootstrapped 95\% confidence intervals. Bold values refer to the best average performance for each model and experiment. For the hospital experiment we report the current performance after training on $n$ hospitals for $n \in \{7,14,21,28,35\}$ in addition to final performance (i.e. after all hospitals). Bracketed numbers refer to the number of different hospitals sequentially trained on thus far. }

\label{tab:results}
\end{table*}

We present the results of the Domain Incremental experiments in Table ~\ref{tab:results}. For brevity we show only the results on outcome of 48hr mortality, see Appendix \ref{sec:full-results} for results on other outcomes (ARF, shock). Results show the final average test balanced accuracy across all tasks for each method. Reported values are means over 5 runs from random initialisation, with bootstrapped 95\% confidence intervals.

For the \textsc{Hospital} domain shift experiments we present the average performance on all tasks thus-seen as the number of tasks increases (i.e. as the models encounter an increasing number of hospitals). Figure \ref{fig:stream-results} displays this performance over time graphically (for the training data).

\subsection{Model Architectures}


Models are generally comparable over a small but constant number (40) of training epochs per domain shift, with the exception of Transformers which demonstrated much more volatile performance over repeated runs. 

Highest training efficiency (measured by number of training epochs required to saturate the current task's loss) was achieved by MLP, followed by LSTM. However a higher training efficiency was correlated with faster (and greater) forgetting upon introduction of new tasks (see for example, MLP EWC vs CNN EWC in Figure \ref{fig:exp-results}). We are currently working on introducing an early stopping mechanism to terminate training on each task only once saturation of a given metric has been achieved (as opposed to a fixed number of epochs) to enable a fairer comparison of methods.

\subsection{Continual Learning strategies} 

\textbf{Regularization} methods showed superior or comparable performance with replay based methods across limited number of domain shifts (\textsc{Age}, \textsc{Ward}, and \textsc{Ethnicity (broad)}, Table \ref{tab:results} top), but decreasing performance as the number of tasks grew large. LwF achieved superior performance on the largest amount of experiments, achieving the lowest degrees of forgetting (note the `flatline' shape of the LwF task curves in Figure \ref{fig:exp-results}). For the \textsc{Hospital} domain shift experiments, regularization methods failed to mitigate catastrophic forgetting for $n$ tasks $\geq 5$, performing on par with Naive fine tuning (no statistically significant difference in final performances). Such performance is expected of regularization methods on domain incremental problems, having been observed in toy problems generally \citep{han2021continual, kirkpatrick2017overcoming}, and in recurrent networks specifically \citep{cossu2021continual}. This is likely due to regularization methods only `delaying the inevitable' when faced with a large number of tasks, as models are `walled off' into shrinking locally optimised regions for parameters.

\textbf{Rehearsal} methods outperformed all other strategies for a large number of domain shifts. This is consistent with class- and domain-incremental results in other benchmarks \citep{lopez2017gradient}. Rehearsal methods all improved with larger storage capacity (Appendix \ref{sec:buffer-size}).

\begin{figure*}[tb!]
\begin{adjustbox}{max width=\textwidth}
	\centering
    	\includegraphics[width=\textwidth]{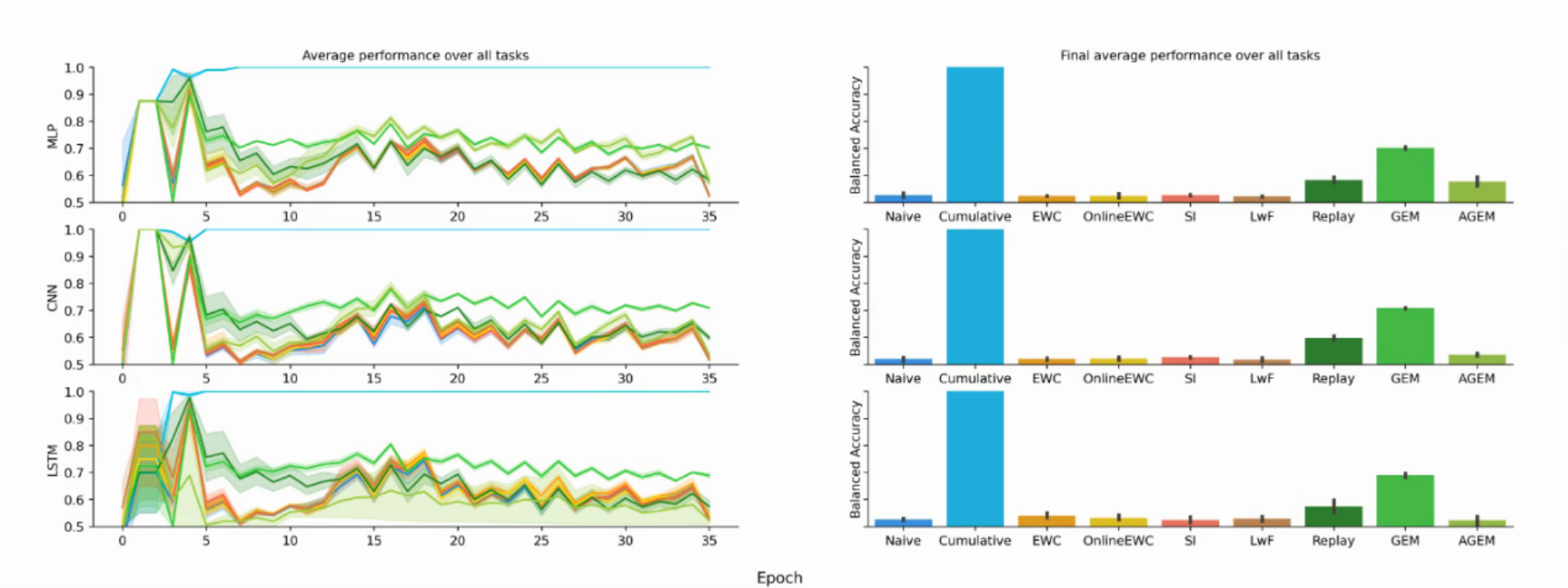}
\end{adjustbox}
\caption{Domain Incremental results for an outcome of mortality (48h) across domain shift of different \textsc{Hospital}. Results show the average training balanced accuracy over all tasks thus encountered, for each model and strategy. Shaded regions (left) and black bars (right) refer to bootstrapped 95\% confidence intervals over 5 runs. Regularization strategies (orange-reds) mitigate catastrophic forgetting to an extent for the first few tasks (hospitals) encountered, but quickly drop to the same performance as the \textsc{naive} baseline (dark blue). \textsc{A-GEM} (lime green) suffers similar behaviour due to averaging of past memory gradients being insufficient to capture the variability in domains. Rehearsal style methods achieve superior performance across the entire range of tasks, with explicit \textsc{Replay} achieving the highest performance in all but one instance. No method achieves comparative performance with \textsc{Cumulative} upper bound (light blue) for $n$ tasks $\geq 5$.}
\label{fig:stream-results}
\end{figure*}

As shown in Figure \ref{fig:stream-results}, regularization methods were generally volatile across a large number of domain shifts, likely corresponding to sets of hospitals more or less similar to the first few encountered. Contrary to this, the rehearsal methods \textsc{A-GEM} and \textsc{GEM} showed relatively stable performance as more hospitals were encountered. This stability in performance over domain shifts demonstrates sustained generalisation as the task population becomes more heterogeneous.



\section{Discussion}
\label{sec:discussion}

Our experiments show that simple deep neural networks trained on rich multi-variate sequential data are also prone to catastrophic forgetting in a domain incremental setting.


We observe that regularization methods are prone to more forgetting than rehearsal based methods across a large sequence of tasks, but for few tasks achieve superior or comparable performance to replay based methods (given a fixed small replay buffer).

In the case of patient health records, data may comprise sensitive patient data and hence sharing between institutions or storage over time may require data sharing agreements and ethical approval. This may be prohibitively time-consuming or infeasible, making rehearsal based methods inapplicable. Data-free rehearsal methods such as generative models overcome this issue, but there is a high computational burden to the learning of accurate generative models for such time-series data.

Future work to be performed:

\begin{itemize}
    \item Implement early stopping mechanism to allow all model/strategies to saturate in current performance before training on new task(s).
    \item Complete supplementary experiments.
    \item Investigate domain shift across different countries / healthcare systems / datasets i.e. :
    \begin{itemize}
        \item MIMIC
        \item eICU
        \item HIRID \citep{yeche2021hirid}
        \item AmsterdamUMCDB \citep{thoral2021sharing}
    \end{itemize} 
    \item investigate MIMIC-IV. The seasonal information of MIMIC-III appears to be too obfuscated, since models do not seriously undergo catastrophic forgetting in this domain. Use annual information preserved in MIMIC-IV for more realistic experiment.
    \item Explore continual learning as a means of bias mitigation (compare CL methods on demographic splits with traditional bias mitigation strategies).
\end{itemize}


\section{Acknowledgements}

Jacob Armstrong is supported by the EPSRC Center for Doctoral Training in Health Data Science (EP/S02428X/1).


\clearpage

\bibliographystyle{unsrtnat}
\bibliography{references}


\clearpage

\onecolumn

\begin{appendices}

\section{Full results}
\label{sec:full-results}

\subsection{Additional outcomes}

Here we present the results for predicting additional outcomes omitted in the main results for brevity, namely the domain incremental experiments on outcomes of ARF (4h) and Shock (4h):

\begin{table*}[ht]
    \centering
    \begin{adjustbox}{max width=\textwidth}
\begin{tabular}{ll|llll|llll}
\multicolumn{10}{c}{\textbf{ARF (4h)}} \\
{} & {} & \multicolumn{4}{c}{\textsc{Age}} & \multicolumn{4}{c}{\textsc{Ethnicity}} \\
{} & {} & {CNN} & {LSTM} & {MLP} & {Transformer} & {CNN} & {LSTM} & {MLP} & {Transformer} \\
\hline
\multirow[c]{2}{*}{Baseline} & Cumulative & 67.9$_{\pm0.7}$ & 64.8$_{\pm1.2}$ & 65.2$_{\pm1.6}$ & 67.2$_{\pm0.7}$ & 66.1$_{\pm0.3}$ & 66.0$_{\pm1.2}$ & 69.5$_{\pm0.3}$ & 67.8$_{\pm1.3}$ \\
 & Naive & 67.2$_{\pm0.3}$ & 66.7$_{\pm0.6}$ & 62.1$_{\pm0.8}$ & 65.8$_{\pm1.0}$ & 69.0$_{\pm0.8}$ & 68.5$_{\pm1.2}$ & 68.3$_{\pm0.5}$ & 66.8$_{\pm1.8}$ \\
 \\
\multirow[c]{4}{*}{Regularization} & EWC & 66.7$_{\pm0.5}$ & \bfseries 66.7$_{\pm1.3}$ & 62.7$_{\pm0.7}$ & 64.2$_{\pm0.9}$ & 68.7$_{\pm0.3}$ & 69.1$_{\pm0.7}$ & 68.3$_{\pm0.7}$ & 66.6$_{\pm1.2}$ \\
 & LwF & 67.3$_{\pm0.1}$ & 66.2$_{\pm0.8}$ & \bfseries 65.4$_{\pm1.4}$ & 65.4$_{\pm1.0}$ & 69.0$_{\pm0.1}$ & \bfseries 69.4$_{\pm0.6}$ & 67.6$_{\pm1.5}$ & \bfseries 67.4$_{\pm1.5}$ \\
 & OnlineEWC & 67.1$_{\pm0.6}$ & 65.2$_{\pm1.1}$ & 62.8$_{\pm0.9}$ & 65.1$_{\pm1.7}$ & 68.4$_{\pm0.8}$ & 68.7$_{\pm0.5}$ & \bfseries 68.5$_{\pm0.4}$ & 65.4$_{\pm1.0}$ \\
 & SI & \bfseries 67.4$_{\pm0.2}$ & 66.5$_{\pm0.9}$ & 63.1$_{\pm0.9}$ & 65.2$_{\pm1.0}$ & \bfseries 69.1$_{\pm0.2}$ & 69.1$_{\pm0.7}$ & 68.1$_{\pm1.1}$ & 66.7$_{\pm2.0}$ \\
 \\
\multirow[c]{3}{*}{Rehearsal} & AGEM & 66.4$_{\pm0.6}$ & 66.2$_{\pm0.6}$ & 59.3$_{\pm0.6}$ & \bfseries 65.6$_{\pm0.9}$ & 68.8$_{\pm0.6}$ & 68.9$_{\pm0.1}$ & 68.2$_{\pm0.7}$ & 64.4$_{\pm5.1}$ \\
 & GEM & 61.3$_{\pm0.3}$ & 59.9$_{\pm0.7}$ & 57.7$_{\pm1.1}$ & 61.9$_{\pm0.4}$ & 68.3$_{\pm0.2}$ & 65.8$_{\pm0.8}$ & 68.3$_{\pm0.3}$ & 66.8$_{\pm1.0}$ \\
 & Replay & 61.3$_{\pm2.0}$ & 62.8$_{\pm0.5}$ & 58.5$_{\pm0.9}$ & 64.1$_{\pm0.5}$ & 67.1$_{\pm1.2}$ & 66.8$_{\pm1.0}$ & 66.9$_{\pm0.8}$ & 65.3$_{\pm1.5}$ \\
\end{tabular}
\end{adjustbox}
\caption{Results for outcome of 4h Acute Respiratory Failure. Similar to the main results on mortality, regularization methods achieve best performance over a limited number of domain shifts. Transformers achieve much more stable performance over the shorter sequence experiments.}
\end{table*}

\begin{table*}[ht]
    \centering
    \begin{adjustbox}{max width=\textwidth}
\begin{tabular}{ll|llll|llll}
\multicolumn{10}{c}{\textbf{Shock (4h)}} \\
{} & {} & \multicolumn{4}{c}{\textsc{Age}} & \multicolumn{4}{c}{\textsc{Ethnicity}} \\
{} & {} & {CNN} & {LSTM} & {MLP} & {Transformer} & {CNN} & {LSTM} & {MLP} & {Transformer} \\
\hline
\multirow[c]{2}{*}{Baseline} & Cumulative & 62.3$_{\pm0.4}$ & 64.3$_{\pm1.4}$ & 65.0$_{\pm0.8}$ & 67.3$_{\pm0.6}$ \\
 & Naive & 65.0$_{\pm0.6}$ & 65.9$_{\pm0.9}$ & 65.3$_{\pm0.3}$ & 64.1$_{\pm1.8}$ \\
 \\
\multirow[c]{4}{*}{Regularization} & EWC & 64.8$_{\pm0.5}$ & \bfseries 67.5$_{\pm0.7}$ & \bfseries 66.4$_{\pm0.5}$ & 63.3$_{\pm2.5}$ \\
 & LwF & 65.1$_{\pm0.5}$ & 66.8$_{\pm0.3}$ & 65.3$_{\pm1.0}$ & \bfseries 65.6$_{\pm0.5}$ \\
 & OnlineEWC & 64.8$_{\pm0.6}$ & 67.1$_{\pm0.7}$ & 65.8$_{\pm0.8}$ & 63.1$_{\pm1.5}$ \\
 & SI & \bfseries 65.3$_{\pm0.4}$ & 67.5$_{\pm0.6}$ & 65.1$_{\pm0.9}$ & 63.5$_{\pm1.4}$ \\
 \\
\multirow[c]{3}{*}{Rehearsal} & AGEM & 62.3$_{\pm0.7}$ & 64.7$_{\pm1.3}$ & 65.1$_{\pm0.7}$ & 63.2$_{\pm0.4}$ \\
 & GEM & 61.7$_{\pm0.5}$ & 61.8$_{\pm1.1}$ & 62.5$_{\pm0.6}$ & 63.0$_{\pm1.1}$ \\
 & Replay & 61.0$_{\pm0.5}$ & 60.7$_{\pm0.7}$ & 62.2$_{\pm0.9}$ & 63.2$_{\pm0.9}$ \\
\end{tabular}
\end{adjustbox}
\caption{Results for outcome of 4h Shock.}
\end{table*}

In contrast to other outcomes, prediction of shock (4h) shows little variation between the naive baseline, continual learning methods, and cumulative upper bound. This may be due to shock presenting similarly across domain shifts

\subsection{Additional sequential models}
\label{sec:additional-sequential-models}

\begin{tcolorbox}[colback=red!5,colframe=red!75!black,title=Work in progress \faWrench ]

Here we evaluate a number of other sequential model architectures omitted from the main results for brevity. namely, we evaluate an RNN and GRU (in addition to the LSTM of the main results).

\end{tcolorbox}

\clearpage

\section{Model and data specifications}

\subsection{Hyperparameters}
\label{sec:hyperparameters}

Hyperparameter tuning was performed via grid search over the following discrete space (parameter names refer to their \texttt{kwarg} names in the Avalanche implementations \citep{lomonaco2021avalanche}):

\begin{table}[h]
\centering

\begin{adjustbox}{max width=0.9\columnwidth}
\begin{tabular}{ll}
\toprule
Hyperparameter &                                Values \\
\midrule
\texttt{mem\_size}         &                           \{256\} \\
\texttt{patterns\_per\_exp} &                        \{256\} \\
\texttt{sample\_size}      &                   \{256, 512\} \\
\texttt{ewc\_lambda}       &  \{0.001, 0.01, 0.1, 1, 10, 100\} \\
\texttt{si\_lambda}        &  \{0.001, 0.01, 0.1, 1, 10, 100\} \\
\texttt{lambda\_e}         &  \{0.001, 0.01, 0.1, 1, 10, 100\} \\
\texttt{alpha}            &  \{0.001, 0.01, 0.1, 1, 10, 100\} \\
\texttt{temperature}      &        \{0.5, 1.0, 1.5, 2.0, 2.5, 3.0\} \\
\texttt{decay\_factor}     &          \{0.2, 0.4, 0.6, 0.8, 0.9, 1\} \\
\texttt{memory\_strength}  &          \{0.2, 0.4, 0.6, 0.8, 0.9, 1\} \\
\bottomrule
\end{tabular}

\begin{tabular}{llllll}
\toprule
Hyperparameter &  Values &    MLP &    CNN &   LSTM &  Transformer \\
\midrule
hidden\_dim     &       [64, 128, 256] &   \checkmark &   \checkmark &   \checkmark &         \checkmark \\
n\_layers       &               [3, 4] &   \checkmark &   \checkmark &   \checkmark &         \checkmark \\
nonlinearity   &  [relu, tanh*] &   \checkmark &   \checkmark &  \xmark &         \checkmark \\
n\_heads        &         [12, 16, 24] &  \xmark &  \xmark &  \xmark &         \checkmark \\
bidirectional  &        [True, False] &  \xmark &  \xmark &   \checkmark &        \xmark \\
\bottomrule
\end{tabular}

\end{adjustbox}
\caption{Grid for method hyperparameter search for all experiments. Left table refers to strategy specific hyperparameters. Right table refers to model specific hyperparameters. Check marks and crosses detail whether hyperparameters are included in the respective model. *gelu nonlinearity used instead of tanh for the Transformer model.}
\label{tab:hyperparameters-grid}
\end{table}

Tuned parameters for each model (base model and CL strategy) and experiment are listed below:

\begin{table}[hb]
\centering

\begin{adjustbox}{max width=0.9\columnwidth}

\begin{tabular}{lllllll}
\multicolumn{7}{c}{\textsc{Age}} \\

\toprule
            &     & lambda & decay\_factor & temperature & sample\_size & patterns\_per\_exp \\
\midrule
MLP & EWC &    0.1 &              &             &             &                  \\
            & OnlineEWC &   0.01 &          0.9 &             &             &                  \\
            & LwF &    0.1 &              &         1.5 &             &                  \\
            & SI &   0.01 &              &             &             &                  \\
            & Replay &        &              &             &       640.0 &                  \\
            & AGEM &        &              &             &       128.0 &            128.0 \\
            & GEM &        &              &         0.2 &             &            128.0 \\
CNN & EWC &   0.01 &              &             &             &                  \\
            & OnlineEWC &  100.0 &          0.5 &             &             &                  \\
            & LwF &  0.001 &              &         2.0 &             &                  \\
            & SI &    0.1 &              &             &             &                  \\
            & Replay &        &              &             &       128.0 &                  \\
            & AGEM &        &              &             &       128.0 &            128.0 \\
            & GEM &        &              &         0.4 &             &            128.0 \\
LSTM & EWC &   0.01 &              &             &             &                  \\
            & OnlineEWC &   0.01 &          0.5 &             &             &                  \\
            & LwF &    1.0 &              &         3.0 &             &                  \\
            & SI &   0.01 &              &             &             &                  \\
            & Replay &        &              &             &       128.0 &                  \\
            & AGEM &        &              &             &       128.0 &            128.0 \\
            & GEM &        &              &         0.6 &             &            128.0 \\
Transformer & EWC &   10.0 &              &             &             &                  \\
            & OnlineEWC &   0.01 &          0.6 &             &             &                  \\
            & LwF &  0.001 &              &         0.5 &             &                  \\
            & SI &   10.0 &              &             &             &                  \\
            & Replay &        &              &             &       128.0 &                  \\
            & AGEM &        &              &             &       128.0 &            128.0 \\
            & GEM &        &              &         0.7 &             &            128.0 \\
\bottomrule
\end{tabular}

\begin{tabular}{lllllll}
\multicolumn{7}{c}{\textsc{Ethnicity (broad)}} \\
\toprule
            &     & lambda & decay\_factor & temperature & sample\_size & patterns\_per\_exp \\
\midrule
MLP & EWC &    0.1 &              &             &             &                  \\
            & OnlineEWC &  0.001 &          0.2 &             &             &                  \\
            & LwF &    1.0 &              &         0.5 &             &                  \\
            & SI &  100.0 &              &             &             &                  \\
            & Replay &        &              &             &       128.0 &                  \\
            & AGEM &        &              &             &       128.0 &            128.0 \\
            & GEM &        &              &         0.8 &             &            128.0 \\
CNN & EWC &  0.001 &              &             &             &                  \\
            & OnlineEWC &  100.0 &          0.9 &             &             &                  \\
            & LwF &  100.0 &              &         1.5 &             &                  \\
            & SI &  0.001 &              &             &             &                  \\
            & Replay &        &              &             &       128.0 &                  \\
            & AGEM &        &              &             &       128.0 &            128.0 \\
            & GEM &        &              &         0.8 &             &            128.0 \\
LSTM & EWC &  0.001 &              &             &             &                  \\
            & OnlineEWC &    1.0 &          0.2 &             &             &                  \\
            & LwF &    1.0 &              &         1.0 &             &                  \\
            & SI &   10.0 &              &             &             &                  \\
            & Replay &        &              &             &       128.0 &                  \\
            & AGEM &        &              &             &       128.0 &            128.0 \\
            & GEM &        &              &         0.8 &             &            128.0 \\
Transformer & EWC &  100.0 &              &             &             &                  \\
            & OnlineEWC &   10.0 &          0.8 &             &             &                  \\
            & LwF &   10.0 &              &         3.0 &             &                  \\
            & SI &  100.0 &              &             &             &                  \\
            & Replay &        &              &             &       128.0 &                  \\
            & AGEM &        &              &             &       128.0 &            128.0 \\
            & GEM &        &              &         0.2 &             &            128.0 \\
\bottomrule
\end{tabular}

\end{adjustbox}

\vspace{1cm}

\begin{adjustbox}{max width=0.9\columnwidth}

\begin{tabular}{lllllll}
\multicolumn{7}{c}{\textsc{Time (season)}} \\

\toprule
            &     & lambda & decay\_factor & temperature & sample\_size & patterns\_per\_exp \\
\midrule
MLP & EWC &    0.1 &              &             &             &                  \\
            & OnlineEWC &  0.001 &          0.9 &             &             &                  \\
            & LwF &  0.001 &              &         2.5 &             &                  \\
            & SI &   0.01 &              &             &             &                  \\
            & Replay &        &              &             &       128.0 &                  \\
            & AGEM &        &              &             &       128.0 &            128.0 \\
            & GEM &        &              &         0.6 &             &            128.0 \\
CNN & EWC &   10.0 &              &             &             &                  \\
            & OnlineEWC &  0.001 &          0.4 &             &             &                  \\
            & LwF &  0.001 &              &         2.0 &             &                  \\
            & SI &    0.1 &              &             &             &                  \\
            & Replay &        &              &             &       128.0 &                  \\
            & AGEM &        &              &             &       128.0 &            128.0 \\
            & GEM &        &              &         0.4 &             &            128.0 \\
LSTM & EWC &  100.0 &              &             &             &                  \\
            & OnlineEWC &  100.0 &          0.9 &             &             &                  \\
            & LwF &  0.001 &              &         1.0 &             &                  \\
            & SI &   0.01 &              &             &             &                  \\
            & Replay &        &              &             &       128.0 &                  \\
            & AGEM &        &              &             &       128.0 &            128.0 \\
            & GEM &        &              &         0.4 &             &            128.0 \\
Transformer & EWC &  0.001 &              &             &             &                  \\
            & OnlineEWC &  0.001 &          0.2 &             &             &                  \\
            & LwF &    0.1 &              &         1.0 &             &                  \\
            & SI &  0.001 &              &             &             &                  \\
            & Replay &        &              &             &       128.0 &                  \\
            & AGEM &        &              &             &       128.0 &            128.0 \\
            & GEM &        &              &         0.4 &             &            128.0 \\
\bottomrule
\end{tabular}
\begin{tabular}{lllllll}
\multicolumn{7}{c}{\textsc{ICU Ward}} \\

\toprule
            &     & lambda & decay\_factor & temperature & sample\_size & patterns\_per\_exp \\
\midrule
MLP & EWC &    0.1 &              &             &             &                  \\
            & OnlineEWC &  0.001 &          0.9 &             &             &                  \\
            & LwF &   10.0 &              &         1.5 &             &                  \\
            & SI &  0.001 &              &             &             &                  \\
            & Replay &        &              &             &      1280.0 &                  \\
            & AGEM &        &              &             &       512.0 &            256.0 \\
            & GEM &        &              &         0.6 &             &            256.0 \\
CNN & EWC &  100.0 &              &             &             &                  \\
            & OnlineEWC &    1.0 &          0.8 &             &             &                  \\
            & LwF &  100.0 &              &         2.0 &             &                  \\
            & SI &  100.0 &              &             &             &                  \\
            & Replay &        &              &             &      1280.0 &                  \\
            & AGEM &        &              &             &       512.0 &            256.0 \\
            & GEM &        &              &         0.8 &             &            256.0 \\
LSTM & EWC &   0.01 &              &             &             &                  \\
            & OnlineEWC &    1.0 &          0.4 &             &             &                  \\
            & LwF &   10.0 &              &         3.0 &             &                  \\
            & SI &   0.01 &              &             &             &                  \\
            & Replay &        &              &             &      1280.0 &                  \\
            & AGEM &        &              &             &       512.0 &            256.0 \\
            & GEM &        &              &         0.8 &             &            256.0 \\
Transformer & EWC &    0.1 &              &             &             &                  \\
            & OnlineEWC &   0.01 &          0.8 &             &             &                  \\
            & LwF &    0.1 &              &         0.5 &             &                  \\
            & SI &  0.001 &              &             &             &                  \\
            & Replay &        &              &             &      1280.0 &                  \\
            & AGEM &        &              &             &       256.0 &            256.0 \\
            & GEM &        &              &         1.0 &             &            256.0 \\
\bottomrule
\end{tabular}

\end{adjustbox}

\caption{Tuned hyperparameters for main experiments (outcome of \textsc{Mortality (48h)}).}
\label{tab:hyperparameters}
\end{table}

\clearpage

\def\mybar#1#2{
  #1 & #2 {\color{green}\rule{\dimexpr1cm*(#1-#2)/#1}{8pt}}{\color{red}\rule{\dimexpr1cm*#2/#1}{8pt}}}

\subsection{Training partitions}
Total number and number of positive samples in each train/validation/test split for each experiment:

\begin{table}[ht]
\begin{adjustbox}{max width=\columnwidth}
\begin{tabular}{lrrrrrrrrrrrr}
\multicolumn{13}{c}{\textsc{Age}} \\
\toprule
task & \multicolumn{2}{l}{0} & \multicolumn{2}{l}{1} & \multicolumn{2}{l}{2} & \multicolumn{2}{l}{3} & \multicolumn{2}{l}{4} & \multicolumn{2}{l}{5} \\
{} &  Total & Outcome &  Total & Outcome &  Total & Outcome &  Total & Outcome & Total & Outcome & Total & Outcome \\
partition &        &         &        &         &        &         &        &         &       &         &       &         \\
\midrule
train     &  \mybar{10794}{760} &  \mybar{10528}{998} &  \mybar{10365}{1191} &  \mybar{11245}{1407} &  \mybar{9385}{1476} &  \mybar{1798}{309} \\
val       &   \mybar{2308}{185} &   \mybar{2248}{236} &   \mybar{2217}{221} &   \mybar{2402}{300} &  \mybar{2012}{331} &   \mybar{372}{61} \\
test      &   \mybar{2273}{173} &   \mybar{2214}{229} &   \mybar{2179}{259} &   \mybar{2348}{329} &  \mybar{1964}{329} &   \mybar{395}{65} \\
\bottomrule
\\
\end{tabular}
\end{adjustbox}

\begin{adjustbox}{max width=\columnwidth}
\begin{tabular}{lrrrrrrrrrr}
\multicolumn{11}{c}{\textsc{Ethnicity}} \\
\toprule
task & \multicolumn{2}{l}{0} & \multicolumn{2}{l}{1} & \multicolumn{2}{l}{2} & \multicolumn{2}{l}{3} & \multicolumn{2}{l}{4} \\
{} & Total & Outcome & Total & Outcome &  Total & Outcome & Total & Outcome & Total & Outcome \\
partition &       &         &       &         &        &         &       &         &       &         \\
\midrule
train     &  \mybar{6394}{705} &   \mybar{912}{101} &  \mybar{41380}{4722} &  \mybar{2014}{270} &  \mybar{2433}{301} \\
val       &  \mybar{1386}{149} &   \mybar{172}{26} &   \mybar{8880}{1019} &   \mybar{405}{55} &   \mybar{500}{45} \\
test      &  \mybar{1303}{139} &   \mybar{186}{25} &   \mybar{8931}{1060} &   \mybar{434}{61} &   \mybar{521}{57} \\
\bottomrule
\\
\end{tabular}
\end{adjustbox}

\begin{adjustbox}{max width=\columnwidth}
\begin{tabular}{lrrrrrrrrrr}
\multicolumn{11}{c}{\textsc{Ward}} \\
\toprule
task & \multicolumn{2}{l}{0} & \multicolumn{2}{l}{1} & \multicolumn{2}{l}{2} & \multicolumn{2}{l}{3} & \multicolumn{2}{l}{4} \\
{} & Total & Outcome & Total & Outcome & Total & Outcome & Total & Outcome & Total & Outcome \\
partition &       &         &       &         &       &         &       &         &       &         \\
\midrule
train     &   \mybar{771}{92} &   \mybar{762}{23} &  \mybar{2654}{400} &  \mybar{1108}{143} &   \mybar{740}{97} \\
val       &   \mybar{166}{21} &   \mybar{169}{5} &   \mybar{553}{87} &   \mybar{230}{20} &   \mybar{160}{11} \\
test      &   \mybar{151}{16} &   \mybar{189}{8} &   \mybar{540}{79} &   \mybar{225}{19} &   \mybar{159}{10} \\
\bottomrule
\\
\end{tabular}
\end{adjustbox}

\begin{adjustbox}{max width=\columnwidth}
\begin{tabular}{lrrrrrrrr}
\multicolumn{9}{c}{\textsc{Time (Season)}} \\
\toprule
task & \multicolumn{2}{l}{0} & \multicolumn{2}{l}{1} & \multicolumn{2}{l}{2} & \multicolumn{2}{l}{3} \\
{} & Total & Outcome & Total & Outcome & Total & Outcome & Total & Outcome \\
partition &       &         &       &         &       &         &       &         \\
\midrule
train     &  \mybar{1428}{186} &  \mybar{1551}{189} &  \mybar{1510}{178} &  \mybar{1546}{202} \\
val       &   \mybar{322}{31} &   \mybar{309}{43} &   \mybar{342}{35} &   \mybar{305}{35} \\
test      &   \mybar{319}{35} &   \mybar{293}{37} &   \mybar{338}{27} &   \mybar{314}{33} \\
\bottomrule \\
\end{tabular}
\end{adjustbox}
\caption{Train, test, validation and outcome breakdowns for 48h mortality. Red and green bars represent proportion of positive and negative outcomes respectively per partition per task. Hospital splits have been omitted due to space constraints.}
\end{table}

\clearpage

\subsection{Domain splits}

\begin{table}[ht]
\begin{adjustbox}{max width=\columnwidth}

\begin{tabular}{lllllllllllll}
\toprule
     &           &        MICU &        SICU &        CSRU &       TSICU &         CCU &   Neuro ICU & Med-Surg ICU &       CSICU &       CTICU & Cardiac ICU &   CCU-CTICU \\
Dataset & Outcome &             &             &             &             &             &             &              &             &             &             &             \\
\midrule
mimic3 & mortality (48h) &  \checkmark &  \checkmark &  \checkmark &  \checkmark &  \checkmark &             &              &             &             &             &             \\
     & ARF (4h) &  \checkmark &  \checkmark &  \checkmark &  \checkmark &  \checkmark &             &              &             &             &             &             \\
     & Shock (4h) &  \checkmark &  \checkmark &  \checkmark &  \checkmark &  \checkmark &             &              &             &             &             &             \\
     & ARF (12h) &  \checkmark &  \checkmark &  \checkmark &  \checkmark &  \checkmark &             &              &             &             &             &             \\
     & Shock (12h) &  \checkmark &  \checkmark &  \checkmark &  \checkmark &  \checkmark &             &              &             &             &             &             \\
eicu & mortality (48h) &  \checkmark &  \checkmark &             &             &             &  \checkmark &   \checkmark &  \checkmark &  \checkmark &  \checkmark &  \checkmark \\
     & ARF (4h) &  \checkmark &  \checkmark &             &             &             &  \checkmark &   \checkmark &  \checkmark &  \checkmark &  \checkmark &  \checkmark \\
     & Shock (4h) &  \checkmark &  \checkmark &             &             &             &  \checkmark &   \checkmark &  \checkmark &  \checkmark &  \checkmark &  \checkmark \\
     & ARF (12h) &  \checkmark &  \checkmark &             &             &             &  \checkmark &   \checkmark &  \checkmark &  \checkmark &  \checkmark &  \checkmark \\
     & Shock (12h) &  \checkmark &  \checkmark &             &             &             &  \checkmark &   \checkmark &  \checkmark &  \checkmark &  \checkmark &  \checkmark \\
\bottomrule
\end{tabular}

\end{adjustbox}
\caption{Domain shifts exhibited for the subset of patients in each outcome dataset.}
\label{tab:partition}
\end{table}

\begin{table}[ht]
\begin{adjustbox}{max width=\columnwidth}

\begin{tabular}{llllllll}
\toprule
     &           &    Hispanic &       Asian & Other/Unknown &   Caucasian & African American & Native American \\
Dataset & Outcome &             &             &               &             &                  &                 \\
\midrule
eicu & mortality 48h &  \checkmark &  \checkmark &    \checkmark &  \checkmark &       \checkmark &                 \\
     & ARF 4h &  \checkmark &  \checkmark &    \checkmark &  \checkmark &       \checkmark &      \checkmark \\
     & Shock 4h &  \checkmark &  \checkmark &    \checkmark &  \checkmark &       \checkmark &      \checkmark \\
     & ARF 12h &  \checkmark &  \checkmark &    \checkmark &  \checkmark &       \checkmark &                 \\
     & Shock 12h &  \checkmark &  \checkmark &    \checkmark &  \checkmark &       \checkmark &                 \\
\bottomrule
\end{tabular}

\end{adjustbox}
\caption{Domain shifts exhibited for the subset of patients in each outcome dataset.}
\label{tab:partition_2}
\end{table}

\newcommand*\rot{\rotatebox{90}}

\begin{table}[ht]
\begin{adjustbox}{max width=\columnwidth}

\begin{tabular}{llp{0.6cm}p{0.6cm}p{0.6cm}p{0.6cm}p{0.6cm}p{0.6cm}p{0.6cm}p{0.6cm}p{0.6cm}p{0.6cm}p{0.6cm}p{0.6cm}p{0.6cm}p{0.6cm}p{0.6cm}p{0.6cm}p{0.6cm}p{0.6cm}p{0.6cm}p{0.6cm}p{0.6cm}p{0.6cm}p{0.6cm}p{0.6cm}p{0.6cm}}
\toprule
       &           & \rot{UNKNOWN/NOT SPECIFIED} & \rot{ASIAN - CHINESE} & \rot{      WHITE} & \rot{HISPANIC OR LATINO} & \rot{WHITE - OTHER EUROPEAN} & \rot{ASIAN - VIETNAMESE} & \rot{ PORTUGUESE} & \rot{HISPANIC/LATINO - DOMINICAN} & \rot{BLACK/AFRICAN} & \rot{UNABLE TO OBTAIN} & \rot{PATIENT DECLINED TO ANSWER} & \rot{ASIAN - ASIAN INDIAN} & \rot{HISPANIC/LATINO - PUERTO RICAN} & \rot{WHITE - RUSSIAN} & \rot{      ASIAN} & \rot{MULTI RACE ETHNICITY} & \rot{BLACK/CAPE VERDEAN} & \rot{BLACK/HAITIAN} & \rot{      OTHER} & \rot{WHITE - BRAZILIAN} & \rot{BLACK/AFRICAN AMERICAN} & \rot{MIDDLE EASTERN} & \rot{ASIAN - FILIPINO} & \rot{HISPANIC/LATINO - GUATEMALAN} & \rot{WHITE - EASTERN EUROPEAN} \\
Dataset & Outcome &                       &                 &             &                    &                        &                    &             &                             &               &                  &                            &                      &                                &                 &             &                      &                    &               &             &                   &                        &                &                  &                              &                          \\
\midrule
mimic3 & mortality 48h &            \checkmark &      \checkmark &  \checkmark &         \checkmark &             \checkmark &         \checkmark &  \checkmark &                  \checkmark &    \checkmark &       \checkmark &                 \checkmark &           \checkmark &                     \checkmark &      \checkmark &  \checkmark &           \checkmark &         \checkmark &    \checkmark &  \checkmark &        \checkmark &             \checkmark &     \checkmark &                  &                              &                          \\
       & ARF 4h &            \checkmark &      \checkmark &  \checkmark &         \checkmark &             \checkmark &         \checkmark &  \checkmark &                  \checkmark &    \checkmark &       \checkmark &                 \checkmark &           \checkmark &                     \checkmark &      \checkmark &  \checkmark &           \checkmark &         \checkmark &    \checkmark &  \checkmark &        \checkmark &             \checkmark &     \checkmark &       \checkmark &                   \checkmark &               \checkmark \\
       & Shock 4h &            \checkmark &      \checkmark &  \checkmark &         \checkmark &             \checkmark &         \checkmark &  \checkmark &                  \checkmark &    \checkmark &       \checkmark &                 \checkmark &           \checkmark &                     \checkmark &      \checkmark &  \checkmark &           \checkmark &         \checkmark &    \checkmark &  \checkmark &        \checkmark &             \checkmark &     \checkmark &       \checkmark &                   \checkmark &                          \\
       & ARF 12h &            \checkmark &      \checkmark &  \checkmark &         \checkmark &             \checkmark &         \checkmark &  \checkmark &                  \checkmark &    \checkmark &       \checkmark &                 \checkmark &           \checkmark &                     \checkmark &      \checkmark &  \checkmark &           \checkmark &         \checkmark &    \checkmark &  \checkmark &        \checkmark &             \checkmark &                &       \checkmark &                   \checkmark &               \checkmark \\
       & Shock 12h &            \checkmark &      \checkmark &  \checkmark &         \checkmark &             \checkmark &         \checkmark &  \checkmark &                  \checkmark &    \checkmark &       \checkmark &                 \checkmark &           \checkmark &                     \checkmark &      \checkmark &  \checkmark &           \checkmark &         \checkmark &    \checkmark &  \checkmark &        \checkmark &             \checkmark &     \checkmark &       \checkmark &                   \checkmark &                          \\
\bottomrule
\end{tabular}

\end{adjustbox}
\caption{Domain shifts exhibited for the subset of patients in each outcome dataset.}
\label{tab:partition_3}
\end{table}


\clearpage

\begin{tcolorbox}[colback=red!5,colframe=red!75!black,title=Work in progress \faWrench]

\section{Additional experiments}

\subsection{Generic regularization}
\label{sec:dropout}

Here we investigate the effect of architecture-agnostic regularization methods on forgetting. We investigate:

\begin{itemize}
    \item dropout $p \in {0.2,0.4,0.6,0.8}$
    \item L2 regularization
    \item SGD momentum {$\in {0.9,0.8,0.7,0.6,0.4,0.2}$}
    \item training batch size {$n \in {16,32,64,128}$}
\end{itemize}

Results:

[...]


\subsection{Sequence length}

Here we evaluate the effect of sequence length on forgetting. We subsample the data stream at a more granular level of 1-hourly, 2-hourly, 4-hourly, 8-hourly, 12-hourly, and 24-hourly. We opted for sub sampling since truncating the datastreams may unfairly bias performance towards larger streams as most pertinent information (in terms of clinically observed changes to vital signs) to patient deterioration occurs closer to the deterioration event \citep{brekke2019value}.



\subsection{Replay buffer size}
\label{sec:buffer-size}

Here we evaluate the rehearsal methods with an increasing storage buffer (from 10\% of the training examples incrementing to full memory i.e. Cumulative strategy).

\subsection{Curricula}
\label{sec:curriculum}

Here we evaluate the comparative performance of models given different curriculum orderings of their tasks. We consider random, correlated, reverse-correlated orderings for the (i) age, (ii) region, and (iii) time experiments.

\subsection{Reduced feature set}
\label{sec:vitals}

Here we evaluate the models on a reduced feature set consisting only of routinely recorded vital signs (as well as static demographic information).

\end{tcolorbox}

\end{appendices}

\end{document}